\documentclass{article}

\usepackage{arxiv}

\usepackage[utf8]{inputenc} % allow utf-8 input
\usepackage[T1]{fontenc}    % use 8-bit T1 fonts
\usepackage{hyperref}       % hyperlinks
\usepackage{url}            % simple URL typesetting
\usepackage{booktabs}       % professional-quality tables
\usepackage{amsfonts}       % blackboard math symbols
\usepackage{nicefrac}       % compact symbols for 1/2, etc.
\usepackage{microtype}      % microtypography
\usepackage{lipsum}
\usepackage{graphicx}
\usepackage{amsmath}
\usepackage{algorithmic}

\graphicspath{ {./figures/} }

\title{Surrogate-Based Black-Box Optimization Method \\for Costly Molecular Properties}
\date{2021}

\author{
Jules Leguy \\
Univ Angers, LERIA,\\
SFR MATHSTIC,\\
F-49000 Angers, France\\
\texttt{jules.leguy@univ-angers.fr} \\
   \And
Thomas Cauchy \\
Univ Angers, CNRS, \\ 
MOLTECH-ANJOU, \\
SFR MATRIX, \\
F-49000 Angers, France\\
\texttt{thomas.cauchy@univ-angers.fr} \\
  \And
Béatrice Duval \\
Univ Angers, LERIA,\\
SFR MATHSTIC,\\
F-49000 Angers, France\\
\texttt{beatrice.duval@univ-angers.fr} \\
  \And
Benoit Da Mota \\
Univ Angers, LERIA,\\
SFR MATHSTIC,\\
F-49000 Angers, France\\
\texttt{benoit.damota@univ-angers.fr} \\
}

\usepackage{tikz}
\usepackage{textcomp}
\usepackage{hyperref}
\usepackage{lipsum}

\newcommand\copyrighttext{%
  \footnotesize \textcopyright 2021 IEEE. Personal use of this material is permitted. Permission from IEEE must be obtained for all other uses, in any current or future media, including reprinting/republishing this material for advertising or promotional purposes, creating new collective works, for resale or redistribution to servers or lists, or reuse of any copyrighted component of this work in other works}
\newcommand\copyrightnotice{%
\begin{tikzpicture}[remember picture,overlay]
\node[anchor=south,yshift=10pt] at (current page.south) {\fbox{\parbox{\dimexpr\textwidth-\fboxsep-\fboxrule\relax}{\copyrighttext}}};
\end{tikzpicture}%
}

\begin{document}

\maketitle
\copyrightnotice
\begin{abstract}
AI-assisted molecular optimization is a very active research field as it is expected to provide the next-generation drugs and molecular materials. An important difficulty is that the properties to be optimized rely on costly evaluations. Machine learning methods are investigated with success to predict these properties, but show generalization issues on less known areas of the chemical space. We propose here a surrogate-based black box optimization method, to tackle jointly the optimization and machine learning problems. It consists in optimizing the expected improvement of the surrogate of a molecular property using an evolutionary algorithm. The surrogate is defined as a Gaussian Process Regression (GPR) model, learned on a relevant area of the search space with respect to the property to be optimized. We show that our approach can successfully optimize a costly property of interest much faster than a purely metaheuristic approach.
\end{abstract}

\section{Introduction}
% no \IEEEPARstart

% You must have at least 2 lines in the paragraph with the drop letter
% (should never be an issue)

In chemistry, molecules are made up of atoms (vertices), bonded by electronic interactions of various strengths (edges). The possible combinations lead to an enormous space of molecular graphs, in which chemists look for solutions with desired properties.

There is currently a very active research for effective strategies to explore the chemical space and optimize molecular properties~\cite{yang_review_AI, elton_deep_2019}. Methods based on deep learning look promising for drug chemistry. For less explored problems such as the chemistry of molecular materials, evolutionary algorithms have shown competitive and interpretable results~\cite{leguy_evomol_2020}.
 
When we are looking for molecules for materials, the target properties depend on the electrons. Their precise estimation require costly quantum mechanical (QM) calculations. Over the past decade, many work has focused on using machine learning methods to predict these costly electronic properties. Kernel ridge regression models were extensively used, as well as neural networks models~\cite{von_lilienfeld_exploring_2020}. The latter yield state of the art performances when many training points are available. Other models were also investigated, such as Gaussian process regression (GPR)~\cite{bartok_machine_2017, musil_fast_2019}.

Such machine learning methods depend on the size and quality of the dataset for good performances in generalization. The most widely used quantum chemistry dataset for small organic molecules, QM9, has been shown to lack chemical diversity~\cite{QM9, glavatskikh_datasets_2019}. In addition, the examples in the training set with optimized electronic properties will be rare. 

An alternative is to build a training dataset that is relevant for a given property. By selecting samples that satisfy some query strategy, active learning methods can provide models with better generalization performances for a given dataset size~\cite{gubaev_machine_2018}. In case of costly properties, this also reduces the cost of building a dataset from scratch.

Another approach, used in this article, is to combine a machine learning model with an optimization method, using the surrogate-based black-box optimization framework~\cite{vu_surrogate-based_2017}. It is suited for costly and non differentiable objective function, such as QM calculations. The optimization and learning procedures are performed jointly, so that the surrogate of the objective function is learned on a relevant space with respect to the property to be optimized. In this context, the quality of the surrogate does not only depend on its prediction performances on large datasets. It should ideally be efficient even when few data points are available, and its error should decrease fast when new points are selected. It must generally allow for a measure that will be used to jointly guide the optimization, such as an uncertainty estimation. And finally, its cost must be negligible compared to the objective function. Tsuda  \emph{et al.} have proposed a number of black-box approaches to find solutions in the molecular space~\cite{terayama_black-box_2021}. In particular, they have developed a Bayesian optimization method that can efficiently screen solutions in a dataset of candidates~\cite{ueno_combo_2016}. 

In this article, we propose a black-box optimization (BBO) method to tackle the optimization of costly molecular properties. It is based on a GPR model that estimates the values of properties at lower cost than QM computation for molecular space exploration. The model is re-trained regularly as selected points are submitted to exact computation. In order to search for new candidates, an evolutionary algorithm is used to optimize the expected improvement (EI) of the surrogate model on the space of molecular graphs. Thus, our method selects candidates in the space of solutions rather than selecting candidates in a predefined dataset. As we wish to propose a method that can be applied to problems with very little data, it is assessed with minimal prior knowledge. In fact, the starting dataset we use in our experiments contains only a single very simple molecule (methane).

At the best of our knowledge, our method is the first that is designed to optimize costly properties in the molecular space with a surrogate-based BBO approach. The contributions of this article also include the proposal and study of a molecular descriptor that does not require geometrical information, for computational efficiency.

In section~\ref{sect_preliminaries}, we present the general concepts for surrogate-based BBO and molecular representation. In section~\ref{sect_methods}, we detail our optimization method. In section~\ref{sect_surrogate}, we perform a data-efficiency study of two machine learning models with different compromises between cost and accuracy. In section~\ref{sect_BBO}, we introduce the surrogate models into the BBO framework to study their optimization performances. We also show that our approach is more efficient than an evolutionary algorithm with state-of-the-art performances on molecular optimization benchmarks. Finally, we conclude and expose the perspectives of our work in section~\ref{sect_conclusion}.

\section{Preliminaries}

\label{sect_preliminaries}

\subsection{Surrogate-based black-box optimization}

Black-box optimization (BBO) is the field of study of the optimization of functions whose derivatives are either non-existent or not available. Gradient-descent techniques can therefore not be used. These functions are called black boxes as they can only be sampled at some selected points. Generally, black-box optimization methods are used and designed for continuous problems, \emph{i.e.} objective functions of the form $f: \mathbb{R}^d \rightarrow \mathbb{R}$. BBO problems arise in different domains, especially when the result of a costly computer simulation must be optimized~\cite{alarie_bbo_applications}. 
%Depending on the properties of the problem, several BBO approaches can be used. We present these approaches thereafter, with a focus on surrogate-based methods. We follow the classification of Vu \emph{et al.}~\cite{vu_surrogate-based_2017}. 
We present thereafter several BBO approaches, following the classification of Vu \emph{et al.}~\cite{vu_surrogate-based_2017}, with a focus on surrogate-based methods.

The most straightforward approach to solve this class of problems is the use of metaheuristic algorithms, that are often able to find high quality approximate solutions on complex problems by using nature-inspired strategies to explore the space of solutions~\cite{boussaid_metaheuristic_survey_2013}. Metaheurstics can be considered as global search methods in that most of them use strategies to escape from local optima. We will focus in this article on evolutionary algorithms, that apply selective pressure on a population of solutions that are randomly modified using perturbation operators such as mutation or crossover. Simulated annealing and particle swarm optimization algorithms can also be mentioned. The main drawback of metaheuristic algorithms is that they require an important number of calls to $f$, and are therefore not best suited for BBO problems with a costly objective function. 

% \subsubsection{Derivative-free optimization} Derivative-free optimization (DFO) is a field of BBO seeking for generalist methods with strong termination and convergence proofs~\cite{audet_introduction_2017}. DFO methods are mainly local and generally assume that derivatives of $f$ exist, although they are unavailable. Among these methods, we can mention the direct search and the trust-region classes of algorithms\cite{rios_derivative-free_review_2013}. Direct search methods perform a local directional\cite{audet_MADS_2006} or simplex-based\cite{nelder_mead_simplex_1965} optimization. Trust-region methods consist in learning a linear or quadratic model of the objective function in a restricted area of the search space. The model is used to search for a optimum of $f$ in the area and the trust-region is adapted at each iteration~\cite{powell_trust_region_1994}.

Surrogate-based optimization aims to solve BBO problems with costly objective functions in a global manner. The main idea is to learn a surrogate function (or metamodel) that is expected to quickly approximate $f$ and that is used to ease the cost of the optimization. This is the approach we will use to address the problem of molecular optimization in this work. Surrogate-based optimization consists in performing the following steps : design of experiments (DOEs), surrogate function learning and optimization of the merit function~\cite{vu_surrogate-based_2017}. DOEs is performed only once at the beginning of the algorithm and consists in selecting a dataset of points that will be used to learn the initial surrogate model. The main procedure consists then in iteratively training the surrogate model on known points, and using the surrogate to select a set of promising points that are evaluated with $f$ and inserted in the dataset. If a stop criterion is reached (for example a budget of calls to $f$) then the procedure is stopped. Otherwise, it goes back to learning the surrogate function using the extended dataset.

\subsubsection{Design of experiments} The points selected for the DOEs are supposed to uniformly cover the search space. A popular heuristic satisfying this property is the Latin hypercube design, that generates the coordinates of the solutions~\cite{mckay_LHS_1979}. This approach is primarily designed for optimization problems in $\mathbb{R}^d$, and cannot be trivially adapted for problems in the space of (molecular) graphs. For the chemical space, active learning methods use statistical design that maximize some optimality criterion, with respect to the machine learning model~\cite{gubaev_machine_2018}. In this case, solutions are not generated but selected from a predefined dataset. In this work, we use a much simpler approach, as the DOEs of our experiments consists only of a single simple molecule.  

\subsubsection{Surrogate learning} The next step consists in learning the surrogate function using the currently available points. The surrogate could be any machine learning method, but some methods are preferred as they can provide additional information that is useful for the following search step. In particular, Gaussian process regression (GPR) models are widely used as they can estimate the uncertainty of their predictions. GPR models predict for any point $x \in \mathbb{R}^d$ a normal distribution whose mean can be interpreted as the prediction and whose standard deviation can be interpreted as the uncertainty (see section~\ref{learning_surrogate}). This method, often also known as Kriging, is the one we will use in this work.

\subsubsection{Merit function} The core of the surrogate-based optimization consists in searching for promising new points using the surrogate function. A straightforward approach is to optimize the value of the surrogate function directly. Assuming the surrogate estimates $f$ accurately, the search will be lead to high-quality points. However, this approach often leads to disappointing results, and can fail to even find a local optimum~\cite{jones_taxonomy_2001}. A number of merit functions to be optimized were proposed to take advantage of the uncertainty information and improve the optimization results. When used to select a subset of a pre-screened set of solutions, the merit function is often referred as a selection criteria. In this article, we will use the expected improvement (EI). It consists in estimating the expected improvement relatively to the best known point, and it performs a compromise between solutions expected to give small improvements with small uncertainty (intensification) and solutions expected to give high improvement with high uncertainty (exploration)~\cite{jones_EI_1998}. The expected improvement is defined for $x \in \mathbb{R}^d$ as $EI(x) = \mathbb{E}[\textrm{max}(Y - f_{\textrm{max}} - \xi, 0)]$ for a maximization problem, with Y the predicted Gaussian distribution at x, and $f_{\textrm{max}}$ the best observed value of f. $\xi$ is an additional parameter supporting the exploration~\cite{lizotte_practical_2008}. It can be computed following eq. (\ref{eqEI}), with $\Phi$ and $\phi$ the cumulative and probability distribution functions of the standard normal function. $\mu(x)$ and $\sigma(x)$ are the parameters of Y.
\begin{equation}\label{eqEI}
\begin{split}
EI(x) &= (\mu(x) - f_{\textrm{max}} - \xi )\Phi(Z) + \sigma(x)\phi(Z)\\
%Z & = \frac{\mu(x) - f_{\textrm{max}} - \xi}{\sigma(x)}
Z & = {(\mu(x) - f_{\textrm{max}} - \xi)} / {\sigma(x)}
\end{split}
\end{equation}

\subsubsection{Obtaining candidate solutions} The merit function aims to select promising candidates that will be evaluated exactly with $f$. A straightforward approach is to use it as a selection criteria to screen a dataset of candidates solutions. If no dataset is available or if one wishes to search for candidates in the complete space of solutions, an optimization method can be used to optimize the merit function. Different methods can be used depending on the features of the problem to be solved. It is sometimes possible to use exact methods such as branch and bound~\cite{jones_EI_1998}. It is also often reasonable to use metaheuristics, as (i) the surrogate function is supposed to be cheap to evaluate so many evaluations should not hinder the performances of the method, and (ii) the surrogate is only an approximation of $f$ that will not necessarily capture its global optima, so reasonable approximate solutions are sufficient~\cite{vu_surrogate-based_2017}. We are especially interested in evolutionary methods, as they allow quite easily to deal with molecular graphs.
% (see section~\ref{sect_evomol}). 
For other applications, the combination of surrogate-based BBO and evolutionary algorithms has shown interesting performances~\cite{emmerich_single-_2006}. 

\subsection{Representing molecules}
\label{mol_repr}

Molecules are objects in a 3 dimensional space. Finding an accurate 3D representation for a given molecule depends on geometrical optimization procedures where the nuclei are arranged in a spatial configuration that minimizes the total energy. This problem can either be tackled by molecular mechanics (MM) or quantum mechanics (QM). In MM, the atoms are described as force fields, \emph{i.e.} simplified functions based on chemical knowledge in order to reproduce typical interatomic distances and angles. MM is relatively fast and outputs the positions of the nuclei in Cartesian coordinates. In this work, we use the MMFF94 force field~\cite{tosco_bringing_MMFF94_2014}. The second approach, QM calculation, is more expensive and more robust. In addition to the Cartesian coordinates, QM computes the electron wave functions which are necessary to compute a number of electronic properties. Among these properties, we have chosen to work on the function containing the most energetic electron, noted HOMO, because it is the electron involved in chemical reactions. In this work, we perform Density Functional Theory (DFT) computations with B3LYP/3-21G* level of theory, which is already an affordable way to perform such QM calculations.

For practical purposes, such as automatic processing and human readability, molecules are often represented as graphs or other equivalent text representations such as SMILES~\cite{weininger_smiles_1988}. Molecular graphs substitute the geometrical information by edges that represent the covalent (strong) interactions between atoms sharing electrons. The edges are labeled according to a discretization of the bond strength, depending on the number of electrons involved. In this work, we only consider the cases of single, double and triple bonds. We also only consider the case of neutral atoms. With $A$ the set of possible atom types (carbon, nitrogen, ...), we formalize a molecular graph as $G = (V, E, f_a, f_b)$, with $V$ the set of vertices (atoms), $E$ the set of edges (covalent bonds), $f_a:V\rightarrow A$ a function labeling the atom types and $f_b:E \rightarrow \{1, 2, 3\}$ a function labeling the bond types.

\subsubsection{Geometrical descriptors}

The representations presented just above are not best suited to be used as a descriptor in a machine learning algorithm predicting a molecular property. They both depend on the ordering of atoms, and the coordinates representation is not invariant to translation nor rotation. Several descriptors were specially designed for the prediction of electronic properties. They are invariant to the ordering of atoms, translation and rotation, and include relevant geometrical information. 

The Coulomb matrix was proposed as such a molecular descriptor in 2012. It encodes the Coulomb repulsion energy between pairs of atoms, and an approximation of the atomic energies in the diagonal~\cite{rupp_fast_CM_2012}. In order to make the descriptor invariant to the ordering of the atoms, a heuristic can be used to sort the rows and columns. 

Smooth Overlap of Atomic Positions (SOAP) was proposed shortly after as a descriptor encoding the local environments of a molecule~\cite{bartok_representing_SOAP_2013, de_comparing_2016}. Atomic densities are represented by Gaussian function centered on all atoms. They are expanded using spherical harmonics and radial basis functions, to obtain a descriptor that is invariant to rotation. A complete molecule can be represented by averaging all local environments.

    Many body tensor representation (MBTR) is a descriptor that encodes geometrical features in a vector that also exhibits the required invariances. It was proposed recently and has shown better performances than both Coulomb matrix and SOAP~\cite{huo_unified_MBTR_2018}. For this reason, MBTR is the geometrical descriptor we will use in our experiments. It is obtained using a set of functions $g_k$, that compute a scalar value from $k$ atoms of the molecule. As we follow the implementation of Himanen \emph{et al.}, we use three base functions~\cite{himanen_dscribe_2020}. $g_1$ corresponds to the atomic number, $g_2$ correspond to the inverse distance and $g_3$ corresponds to the cosine of the angle. These functions are used to create a set of distributions $D_k(x)$ by using Gaussian kernel density estimation, for all combinations of atom types and following eq. (\ref{eq_MBTR_distributions}).

\begin{equation}
\begin{split}
     \textrm{MBTR}^{Z_1}_1(x) &= \sum_l^{|Z_1|}D^l_1(x)\\
     \textrm{MBTR}^{Z_1, Z_2}_2(x) &= \sum_l^{|Z_1|}\sum_m^{|Z_2|}w_2^{l, m}D^{l, m}_2(x)\\
     \textrm{MBTR}^{Z_1, Z_2, Z_3}_3(x) &= \sum_l^{|Z_1|}\sum_m^{|Z_2|}\sum_n^{|Z_3|}w_3^{l, m, n}D^{l, m, n}_3(x)
\end{split}
    \label{eq_MBTR_distributions}
\end{equation}

$D_k(x)$ are normalized Gaussian distributions centered on the corresponding $g_k$ value. The distributions are scaled by a weight that is inversely proportional to the distances between the atoms when $k\ge2$. The distributions are sampled at evenly spaced bins, yielding a tensor of dimension $k+1$. The MBTR output is the concatenation of all the distributions after individually normalizing the values corresponding to each $k$ term so that they have a unitary L2 norm.

\subsubsection{Graph-based descriptors}

Another approach is to use descriptors based only on the molecular graph, such as molecular fingerprints. Molecular fingerprints are built by applying a hash function on structural features, and also exhibit required invariances. MHFP6 is a fingerprint algorithm based on molecular subgraphs named shingles~\cite{probst_probabilistic_2018}. A shingle is defined for a given atom $a$ of a molecular graph and a given radius $r$ as the graph centered on $a$ that contains all atoms and all bonds that are at a distance less than or equal to $r$ from $a$ in number of bonds. MHFP6 is built by extracting all shingles of diameter up to 6 ($r$ = 3) and using a hash function to obtain a sparse descriptor of reasonable size. We emphasize that these features can be computed much faster than geometrical descriptors as they do not require any geometrical optimization. We will propose in this article a descriptor following the same idea as MHFP6, without using a hash function.

\section{Methods}

\label{sect_methods}

\subsection{Black-box optimization of molecular graphs}

To solve the problem of the optimization of molecular graphs with the aim to minimize the costly calls to the objective function, we propose the following surrogate-based BBO method (see algorithm in Figure~\ref{bboalg}). It is based on the optimization of the EI of the surrogate model, using an evolutionary algorithm (EA). This allows generating new solutions in the molecular space, in opposition to selecting solutions from a predefined dataset. The main loop is executed until the stop condition is reached (a threshold on the number of calls to the objective function). At each step, the surrogate model is trained on the currently available data ($D$). Then, $n$ restarts of the evolutionary optimization are performed, using as initial population different subsets of $D$. Each restart provides a solution that is added to $D$ at the end of the step. The use of restarts rather than a single optimization aims to make the most of the surrogate model on various areas of the chemical space, before needing to call the objective function again. The elements of $D'$ are selected randomly from $D$ according to a random selection weighted by their objective values. Thus, all previously found solutions can be used as starting point for evolutionary optimizations, but solutions with high values are favored. In the implementation, we spare the cost of EI computation for solutions of the initial population by assigning their score a zero value, as they are already known in $D$. For each restart, the chosen solution is the one with the highest EI value that has not been chosen in previous restarts for the same step. The solutions of $D$ are given as a tabu list to the evolutionary algorithm so that only unknown solutions can be generated. Our algorithm is meant to be easily parallelizable, as both evolutionary optimizations and costly objective function calculations can be performed in parallel (internal loop).

\begin{figure}[h!]
\begin{algorithmic}
    \STATE \textbf{input}
    \STATE ~~~$D$: dataset of solutions at initialization
    \STATE ~~~$f$: objective function
    \STATE \textbf{parameter}
    \STATE ~~~$n$: number of solutions to be generated at each step
    \WHILE{the stop condition is not reached}
        \STATE train surrogate $s$ on $D$
        \FOR{$i\leftarrow 1$ to $n$ \COMMENT{restarts} }
            \STATE select a subset $D'$ of $D$ \COMMENT{initial population selection}
%            \STATE $c \leftarrow $ best solution from the optimization of the EI of $s$ using the evolutionary algorithm
            \STATE $c \leftarrow \max \textrm{EA}(\textrm{EI}(s), D')$ \COMMENT{evolutionary optimization}
            \STATE $D \leftarrow   D \cup \{~(c, f(c))~\}$
        \ENDFOR
    \ENDWHILE
    \RETURN $D$
\end{algorithmic}
\caption{Surrogate-based black-box optimization of molecular graphs}
\label{bboalg}
\end{figure}

% \begin{figure}
% \begin{algorithmic}
%     \STATE Let $M$ be the set of all molecular graphs and $d$ be the dimension of the descriptor
%     \STATE \textbf{input}
%     \STATE $f:M \rightarrow \mathbb{R}$ the objective function
%     \STATE $f_{\textrm{desc}} : M \rightarrow \mathbb{R}^d$ the function computing the descriptor
%     \STATE $s: \mathbb{R}^d \rightarrow \mathbb{R}$ the surrogate function
%     \STATE $D \subset M$ the dataset of solutions at initialization (design of experiments)
%     \STATE $n$ the number of solutions to be generated at each step
%     \WHILE{the stop condition is not reached}
%         \STATE train $s$ on data $f_{\textrm{desc}}(D), f(D)$ \COMMENT{update surrogate}
%         \FOR{$i\leftarrow 1$ to $n$ \COMMENT{$n$ restarts}}
%             \STATE $D' \leftarrow$ selection of a subset of $D$ \COMMENT{initial population selection}
%             \STATE $c \leftarrow $ best solution from the optimization of the $EI$ of $s$ starting from $D'$
%             \STATE add $c$ to $D$ \COMMENT{update dataset}
%         \ENDFOR
%     \ENDWHILE
%     \RETURN $D$, f($D$) \COMMENT{Returning the dataset and the objective function values}
% \end{algorithmic}
% \caption{Surrogate-based black-box optimization of molecular graphs}
% \label{bboalgbis}
% \end{figure}

\subsection{Learning a surrogate model}
\label{learning_surrogate}

\subsubsection{Model} In our experiments, we use Gaussian process regression (GPR) models as they are well-suited to be integrated in a surrogate-based black-box optimization framework. A Gaussian process (GP) is a distribution over functions that can model any point $x \in \mathbb{R}^d$ as a Gaussian probability distribution~\cite{rasmussen_gaussian_2006}. A GP is defined by a covariance (or kernel) function $k(x, x')$ and a mean function $m(x)$. Gaussian process regression consists in conditioning a prior GP on the observed data points $X$ with corresponding objective function values $\mathbf{y}$, to obtain a posterior GP that can estimate the distributions $\mathbf{f_*}$ of a new set of points $X_*$. The prior is generally set to have a constant zero mean. For regression on noisy data, an additional parameter $\sigma_n^2$ modeling the variance of the noise is also used. The predicted multivariate normal distribution is expressed in eq. (\ref{eq_GP_distrib}). Its parameters can be computed following eq. (\ref{eq_GP_pred}). For any sets of $n$ and $n'$ elements of $\mathbb{R}^d$ $Y$ and $Y'$, the notation $K(Y, Y')$ represents the matrix $n \times n'$ obtained by computing $k$ on all pairs of elements of $Y$ and $Y'$. $\overline{\mathbf{f}}_*$ corresponds to the mean of the distribution and thus the predictions for the elements of $X_*$. The variance of the predictions can be extracted from the diagonal of $\mathbf{\Sigma_{f_*}}$. The noise $\sigma_n^2$ is inserted in the $K_n = K(X, X) + \sigma_n^2I$ term. We encourage the interested reader to consult the book of Rasmussen \emph{et al.} for a thorough explanation~\cite{rasmussen_gaussian_2006}. In this article, we will use the implementation of scikit-learn~\cite{pedregosa_scikit-learn:_2011}. 
\begin{equation}
    \mathbf{f_*}|X, \mathbf{y}, X_* \sim \mathcal{N}(\overline{\mathbf{f}}_*, \mathbf{\Sigma_{f_*}}) \\
    \label{eq_GP_distrib}
\end{equation}
\begin{equation}
    \begin{split}
    \overline{\mathbf{f}}_* &= K(X_*, X) K_n^{-1}\mathbf{y}\\
    \mathbf{\Sigma_{f_*}} &= K(X_*, X_*) - K(X_*, X) K_n^{-1} K(X, X_*)
    \end{split}
    \label{eq_GP_pred}
\end{equation}

\subsubsection{Kernels} We study the performances of two kernels, namely RBF and dot product. They are defined in equation (\ref{eq_rbf_dotproduct}). $d$ corresponds to the Euclidean distance function. Both kernels depend on a parameter $\sigma_s^2$ that scale their values. All their parameters are optimized during GPR computation by maximizing the log-likelihood on the training data.
\begin{equation}
\begin{split}
     k_{\textrm{RBF}}(x, x') &= \sigma_s^2\textrm{exp}\left(-\frac{d(x, x')^2}{2l^2}\right)\\
     k_{\textrm{Dot-product}}(x, x') &= \sigma_s^2(\sigma_0^2 + x\cdot x')
\end{split}
\label{eq_rbf_dotproduct}
\end{equation}

\subsubsection{Descriptors} As neither the molecular graphs nor the Cartesian coordinates of the atoms can be mapped to $\mathbb{R}^d$ with proper invariances, we use MBTR as input of the machine learning models. However, it should be noted that MBTR depends on geometrical data and thus on a geometrical optimization. Although molecular mechanics optimization is much cheaper that quantum mechanics, it might still be too costly in a BBO framework regularly assessing a lot of candidate solutions. We study another descriptor that we design to depend only on molecular graph data. It is inspired from MHFP6 fingerprints, but encodes more local structures. Rather than using $r = 3$ shingles that suffer from combinatorial explosion, we propose here to only use $r = 1$ shingles and to count their occurrences in an integer vector of fixed size. With such small radius, there is no need for a hash function as 2000 seems to be a reasonable vector size~\cite{leguy_scalable_2021}. Less information can be encoded compared to MHFP6 due to the smaller radius, but no collision can occur and the resulting descriptor is more interpretable. All encoded subgraphs (chemical local environments) can be retrieved from the encoding.

\subsection{Molecular graph evolutionary optimization}
\label{sect_evomol}
Our BBO framework requires a metaheuristic algorithm to optimize the expected improvement of the surrogate function. We use EvoMol, an open-source evolutionary algorithm designed to be flexible~\cite{leguy_evomol_2020}. EvoMol was shown to successfully optimize a various set of molecular properties, including the HOMO energy that is our target in this article. It is based on a set of local mutations of the molecular graph, which are randomly selected among those that maintain the validity of the solutions. At each optimization step, a batch of worst individuals are replaced by improving solutions, obtained by mutating the best individuals in fitness value order.

\section{Surrogate model study}

\label{sect_surrogate}

In this section, we evaluate our GPR models as predictors of the target molecular property, independently of the optimization process that will be studied in section~\ref{sect_BBO}. Our objective is to insure that they meet the requirements to be used as the surrogate of the BBO approach. Firstly, the surrogate model must be significantly faster than the original evaluation function. Secondly, it must be sufficiently accurate, especially with little data. These requirements will depend on the molecular descriptor and kernel associated with the model.

We will focus our study on two models, that are built using chosen combinations of descriptors and kernels presented in previous section. The first one combines the shingles count as descriptors and a dot product kernel, denoted GPR(Shingles,~$\cdot$). It is expected to be a very quick estimator of the property.
The second one combines MBTR descriptors and the RBF kernel, denoted GPR(MBTR,~RBF). It is associated with the state of the art prediction performances~\cite{huo_unified_MBTR_2018}. We emphasize that the computational cost of both MBTR and RBF is higher than that of shingles and dot product respectively.

\subsection{Experiments}

We first study the data efficiency, which corresponds to the performance of the models depending on the number of training samples. We use a set of points randomly selected from the QM9 dataset, that is a reference dataset for machine learning of molecular materials properties~\cite{QM9}. QM9 was built by performing DFT calculations on a subset of an enumeration attempt of the chemical space~\cite{GDB17}. Solutions in QM9 contain up to 9 heavy atoms (\emph{i.e.} not counting the hydrogen) among carbon, oxygen, nitrogen and fluorine. The graphical representation of the data efficiency (learning curves) is built by evaluating the models for different training set sizes. We estimate the mean absolute error (MAE) for HOMO energy prediction by performing 10-folds cross validation on a random subset of 10000 solutions.

Then, we estimate the computational cost of both models, to evaluate the additional cost of our BBO setting. We study a state of the optimization process in which the dataset of solutions $D$ (see algorithm in Figure~\ref{bboalg}) contains 1000 solutions, which corresponds in fact at the maximum possible size that may be reached at the end of the experiments we perform in section~\ref{sect_BBO}. The BBO overhead consists firstly in learning the surrogate on this dataset, and secondly in optimizing its expected improvement. We use the times measured for the cross validation experiment as an estimation of the fit time. We consider that the optimization of the merit function is driven by the time necessary to compute the descriptors of the candidate solutions. A single BBO step is equivalent to the evaluation of 1000 candidates (see section~\ref{sect_BBO}). Thus, we measure the time necessary to compute the descriptors on a set of 1000 solutions. This measure includes the time of MM optimization for MBTR descriptors. We also estimate the mean time of DFT calculations that are predicted here, based on a measure of the evaluations that are performed in the experiments of section~\ref{sect_BBO}.

\subsection{Results}

\begin{figure}
    \centering
    \includegraphics[width=0.6\textwidth]{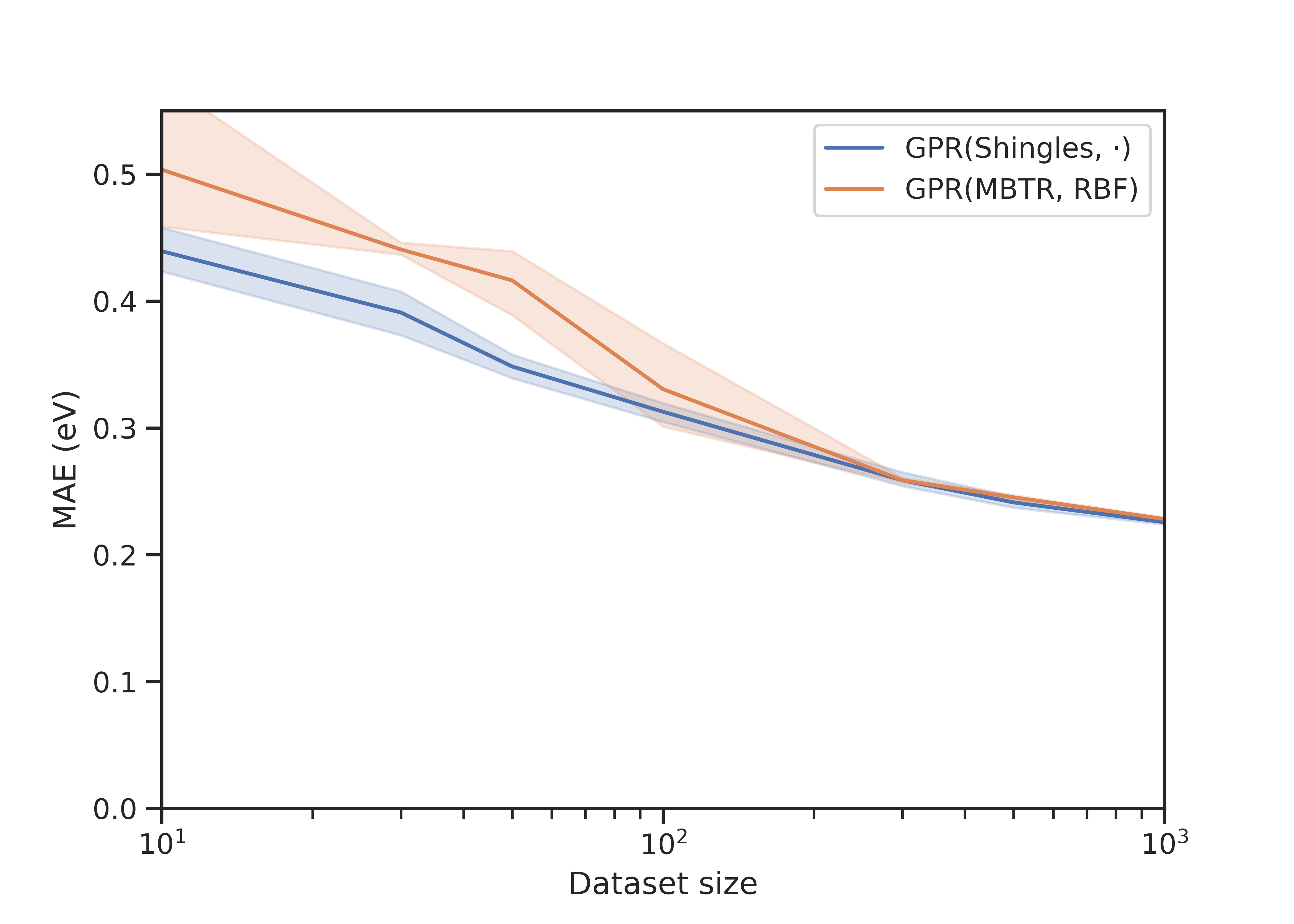}
    \caption{Learning curves of HOMO energies, showing out-of-sample prediction error (mean absolute error) depending on the number of training molecules drawn at random from QM9 dataset for the two proposed surrogate models.
    }
    \label{fig:mae}
\end{figure}

\begin{table}[]
    \centering
    \begin{tabular}{l|rrr}
        \multicolumn{1}{c|}{\textbf{Method}} & \textbf{Model fit} & \textbf{Descriptors} & \textbf{Total}\\ 
        & Worst case & 1000 mol. & per mol.\\
        \hline
        GPR(Shingles,~$\cdot$) & 3.8 & 3.2 & 0.007 \\
        GPR(MBTR,~RBF)         & 26.2 & 102.5 & 0.129 \\
        %\textit{DFT (B3LYP, 3-21G*)} & -& -& \textit{243.1}\\
        \textit{DFT} & -& -& \textit{243.1}\\
    \end{tabular}
    \caption{Computing performances of descriptors and models (in seconds)}
    \label{tab:perfML}
\end{table}

The learning curves are shown in Figure~\ref{fig:mae}. Both models perform comparably when the dataset size increases.
A slight advantage can be observed for GPR(Shingles,~$\cdot$) with very little data. The overall level of error is comparable to the error of the DFT approximation~\cite{zhang_comparison_2007}. Thus, we consider our models to be a reasonable approximation of the HOMO energy, with few data samples. 

In the Table~\ref{tab:perfML}, we observe that the times to fit the model and to compute the descriptors are as expected more important for GPR(MBTR,~RBF). In the last column, we report the estimated time per molecule. We consider for this the time to compute the descriptors for a dataset of 1000 molecules and then to fit the corresponding model (the two previous columns). The prediction time is considered here negligible. The total time is divided by 1000 to account for the time of a single evaluation, to be comparable to a DFT calculation. The GPR(Shingles,~$\cdot$) and GPR(MBTR,~RBF) models are respectively 5 and 3 orders of magnitude faster than DFT. Thus, they meet the requirements to be used as surrogate models in a BBO framework.

\section{Black-box optimization}
\label{sect_BBO}

In this section, we assess our BBO framework on the maximization of the HOMO energy. The machine learning models studied in previous section are integrated as surrogate functions into the BBO framework. They are assessed against an evolutionary algorithm (EA) baseline.

% details
\subsection{Experiments}

We use the algorithm described in Figure~\ref{bboalg} to maximize the HOMO energy, using in distinct experiments the surrogate models GPR(Shingles,~$\cdot$) and GPR(MBTR,~RBF) studied in previous section. These experiments will be referred respectively as BBO(Shingles,~$\cdot$) and BBO(MBTR,~RBF).
The evolutionary optimization of the merit function is performed using 10 restarts, so 10 solutions are submitted to exact computation at each step. For each restart, the population is initialized with 10 solutions drawn from the dataset of known solutions $D$. The maximum population size is set to 300, and 10 solutions are inserted at each merit optimization step. Though the mutations defined in EvoMol are local, they can rapidly have a huge impact on the molecular graph in terms of chemistry. We limit the number of EA optimization steps to 10 so that the resulting solutions are at most 10 mutations away from the known solutions in the initial population. As a mutation is defined as up to 2 perturbations of the molecular graph in EvoMol, this can already lead to huge jumps in the chemical space. As a result, 100 solutions are evaluated for each restart, which overall corresponds to 1000 evaluations of the surrogate at each BBO step. The $\xi$ parameter in the EI definition supporting the exploration of the space of solutions is set to $10^{-2}$.

As stated in the introduction, we use a trivial DOEs consisting only of the methane molecule. This is motivated by the aim to propose a generalist method, that can be used on problems for which few data is available. Furthermore, the methane has actually a very low HOMO energy (-10.6 eV), which makes it a difficult starting point. This allows us to validate our method in an unfavorable setting. We emphasize the fact that for initial optimization steps, the surrogate model has extremely little knowledge about the function to be optimized and that it must acquire it progressively.

Our approach is assessed against an EA baseline. It consists actually in using the same EA as used internally in our BBO method, but used in this situation as a direct optimizer of the HOMO energy value. We remind the reader that this particular method has been previously shown to optimize this property with success~\cite{leguy_evomol_2020}. The parameters are left identical. By using the same algorithm for both internal optimization of the surrogate in the BBO framework, and direct metaheuristic optimization of the property, we aim to observe the actual impact of our framework on the efficiency of the search.

For all experiments, the space of solutions is limited to that of QM9, \emph{i.e.} 9 heavy atoms among carbon, nitrogen, oxygen and fluorine. These criteria define a search that is representative of organic chemistry, with a reasonable cost for DFT calculations. Experiments are stopped when they reach 1000 calls to the DFT calculations, and are performed 10 times so that we can obtain an average performance. Parallelization is disabled and homogeneous computing servers are used in order to allow for fair time measures.

In order to aggregate the results of the different runs, we will use the empirical cumulative distribution functions (ECDF) representation and the expected running time measure (ERT)~\cite{hansen_coco_2021}. Both are based on a set of predefined numerical target values on the objective function. The ECDF represent the proportion of targets that were achieved among the different runs of the same experiments at a given unit of progression. The targets are set between -10 eV and -1 eV with a step size of $10^{-2}$. The ERT corresponds to the mean experimental time to reach a given target. %It is computed as the sum of times before reaching a target, divided by the number of runs that actually reached the target. 
We will study the $-3$ eV target, that corresponds already to a very high HOMO value for the considered chemical space. For both measures we will study the progression in terms of number of calls to the objective function and CPU time.

\subsection{Results}

\begin{figure}
    \centering
    \includegraphics[width=0.6\textwidth]{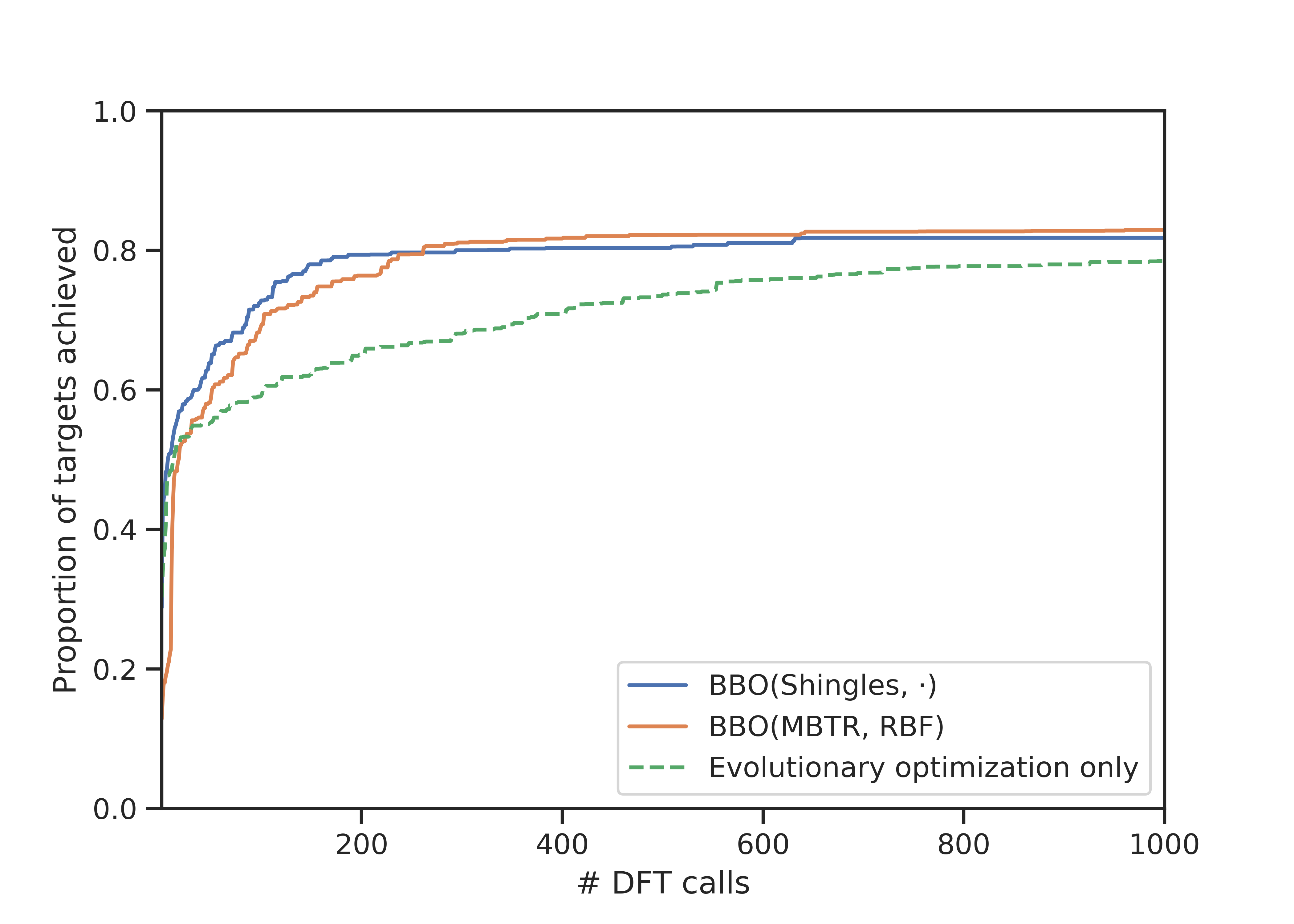}\\
    \includegraphics[width=0.6\textwidth]{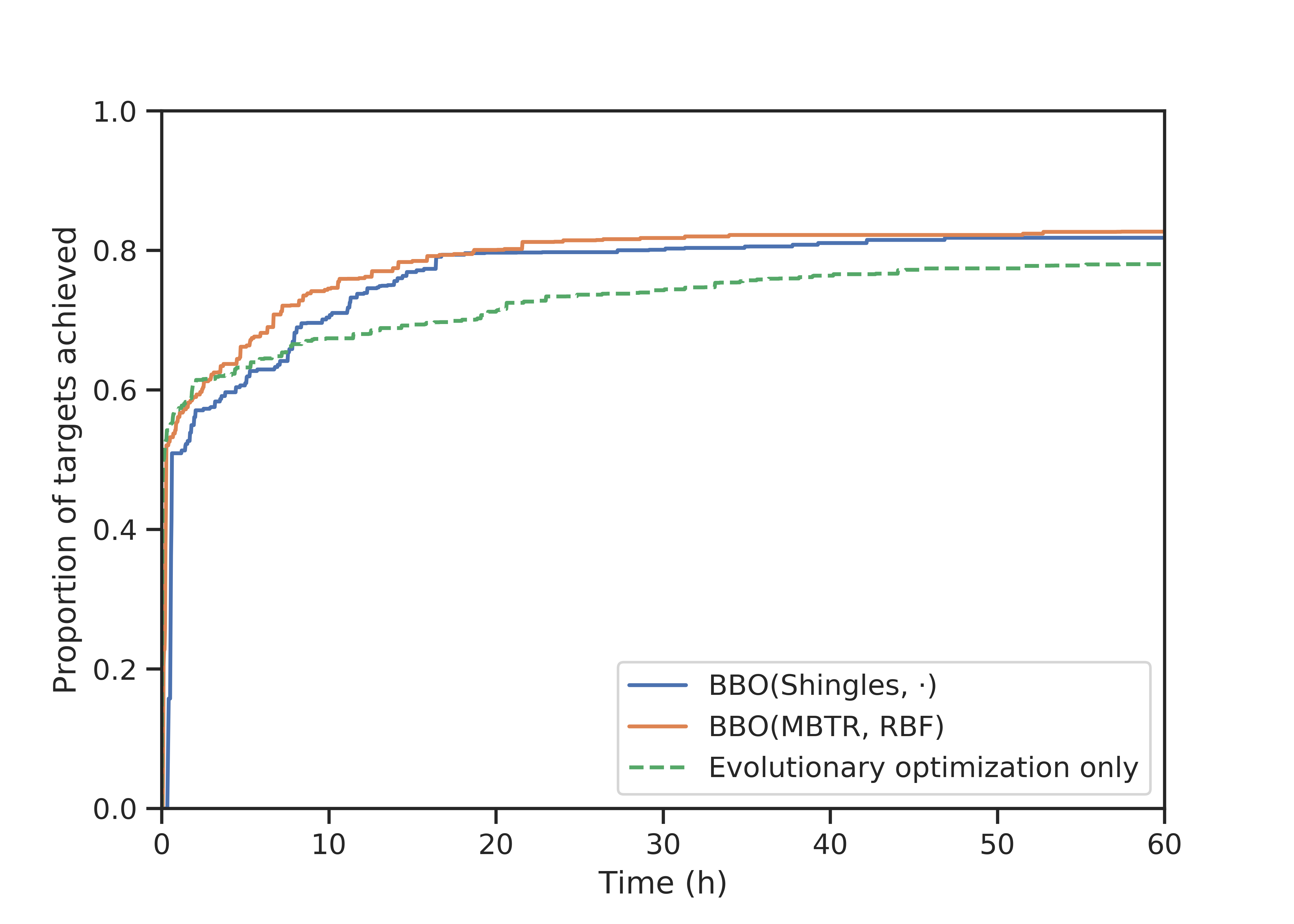}
    \caption{Empirical cumulative distribution functions of the number of DFT calls (top) and CPU time (bottom) for the two proposed surrogate model and the evolutionary algorithm used as a baseline.}
    \label{fig:ecdf}
\end{figure}

\begin{table}[]
    \centering
    \begin{tabular}{l|rr}
         \multicolumn{1}{c|}{\textbf{Method}} & \multicolumn{2}{c}{\textbf{ERT to reach -3 eV}} \\ 
         & \textbf{DFT calls} & \textbf{CPU time (h)}\\ \hline
         BBO(Shingles,~$\cdot$)& 377 & 27.9\\
         BBO(MBTR,~RBF) & 177 & 11.3\\
         Evolutionary optim. & 1186 & 74.2\\
    \end{tabular}
    \caption{Expected running time for HOMO optimization (10 executions)}
    \label{tab_ERT}
\end{table}

It can be observed in the upper part of Figure~\ref{fig:ecdf} that both BBO models are more efficient than the evolutionary optimization in number of calls to the objective function. They seem to have converged after 600 calls, while the EA is still progressing. The results of Table~\ref{tab_ERT} show that the BBO(Shingles,~$\cdot$) model needs twice as much calls as BBO(MBTR,~RBF) to find the high $-3$ eV target. Compared to the EA, they are 3 to almost 7 times more efficient. 

When we study the efficiency in terms of CPU time (lower part of Figure~\ref{fig:ecdf}), both BBO models seem again quite close to each other. However, BBO(MBTR,~RBF) is slightly more efficient during the whole optimization. This is surprising as it depends on more costly components and considering previous results, BBO(Shingles,~$\cdot$) would be expected to be more efficient at least in early stages. It seems that the additional cost of computing MM geometry and using a RBF kernel is worthwhile as the gain in terms of optimization is more important.
We have noticed that in early steps, BBO(MBTR,~RBF) generates small molecules that are fast to evaluate with DFT. The model using shingles directly proposes solutions close to the maximum size of 9 heavy atoms. This explains why early steps are longer for BBO(Shingles,~$\cdot$). The Table~\ref{tab_ERT} shows the importance of the gain of the BBO approach relatively to the evolutionary algorithm. The high -3 eV target can be obtained with our approach in half a day, while it requires more than 3 days for the evolutionary approach.

\section{Conclusion}
\label{sect_conclusion}

In this article, we consider the problem of costly molecular properties optimization using a surrogate-based BBO framework. We propose an approach that consists in using an evolutionary algorithm to optimize the expected improvement of the prediction of a GPR model. The GPR model is used as a surrogate of the costly target property to be optimized. The surrogate allows a fast approximate evaluation of many candidates, that can be generated in the complete space of solutions by the evolutionary algorithm. We also perform a data efficiency study of the GPR model, independently of the optimization procedure. Two molecular descriptors are investigated. Firstly, a state-of-the-art continuous descriptor that depends on molecular geometrical optimization (MBTR). Secondly, a graph-based discrete descriptor of our proposal that can be computed much faster (count of shingles). These descriptors are used in two GPR models that are studied in terms of data-efficiency and computational cost.

The proposed GPR models show satisfying prediction performances. On small training sets, the model using the count of shingles as a descriptor is more accurate. Other simple molecular descriptors that are fast to compute could be expected to yield good performances, such as CUSTODI that was proposed very recently~\cite{fite_custom_2021}. Alternatives to GPR for uncertainty estimation could be to use ensemble methods such as bagging, or Monte Carlo dropout with deep learning models~\cite{gal_dropout_2016}. However, this would still be challenging for deep learning as the surrogate is supposed to learn quickly and with few samples.

Our optimization method is shown to optimize an electronic property very efficiently, starting from minimal knowledge. It is up to 6 times more efficient than a state-of-the-art evolutionary approach. An immediate perspective would be to extend this work on a larger chemical space, with solutions containing more atoms and with additional atom types. These problems are intractable with evolutionary algorithms, as the cost of QM evaluation becomes prohibitive. In this case, it might become necessary to define a more complex design of experiments. Prior knowledge thus introduced could limit the cost of exploring the chemical space in early optimization steps. We believe that an approach based on the diversity of chemical descriptors may be suited for this task.

The surrogate-based black-box optimization approach is very appropriate for costly molecular properties. Moreover, it is a very generalist framework. 
The surrogate function could be optimized by other optimization methods than evolutionary algorithms. Other metaheuristic algorithms, Monte-Carlo tree search of even deep generative methods can be considered. We expect that this field will grow rapidly as it allows for an efficient exploration of the immensity of the chemical space.

% use section* for acknowledgment
\section*{Acknowledgment}

The authors would like to thank Adrien Goëffon for the enriching discussions and the LERIA as well as the CCIPL that have provided the computing resources. This work was supported by the University of Angers and the french Ministry of Education and Research (JL PhD grant).
Sources are available at \url{https://github.com/jules-leguy/BBOMol}

% trigger a \newpage just before the given reference
% number - used to balance the columns on the last page
% adjust value as needed - may need to be readjusted if
% the document is modified later
%\IEEEtriggeratref{8}
% The "triggered" command can be changed if desired:
%\IEEEtriggercmd{\enlargethispage{-5in}}

% references section

% can use a bibliography generated by BibTeX as a .bbl file
% BibTeX documentation can be easily obtained at:
% http://mirror.ctan.org/biblio/bibtex/contrib/doc/
% The IEEEtran BibTeX style support page is at:
% http://www.michaelshell.org/tex/ieeetran/bibtex/
% argument is your BibTeX string definitions and bibliography database(s)
\bibliography{template.bib}

\bibliographystyle{unsrt}  
%\bibliography{references}  %%% Remove comment to use the external .bib file (using bibtex).
%%% and comment out the ``thebibliography'' section.

\end{document}